\documentclass[12pt]{article}
\usepackage[ansinew]{inputenc}
\usepackage[colorlinks=true,linkcolor=blue]{hyperref}
\usepackage[english]{babel}
\usepackage[usenames]{color}

\usepackage{amsbsy}
\usepackage{amsfonts}
\usepackage{amsmath,amscd}
\usepackage{amssymb}
\usepackage{amsthm}
\usepackage{anysize}
\usepackage{bbm}
\usepackage{bookmark}
\usepackage{euscript}
\usepackage{ifsym} 
\usepackage{makeidx}
\usepackage{tocloft}

\newcommand{\Add}{{\rm Ad}}
\newcommand{\B}{\mathbb{B}} 
\newcommand{\ba}{\begin{array}} 
\newcommand{\bea}{\begin{eqnarray}} 
\newcommand{\Cn}{{\rm Cn}}
\newcommand{\CNF}{{\rm CNF}}
\newcommand{\comp}{\raisebox{1.53	mm}{$_{_{\, \, \circ }}$}} 
\newcommand{\con}{\subseteq} 
\newcommand{\De}{\Delta}
\newcommand{\DNF}{{\rm DNF}}
\newcommand{\Ds}{{\rm Ds}}
\newcommand{\DS}{\displaystyle}
\newcommand{\ea}{\end{array}} 
\newcommand{\eea}{\end{eqnarray}} 
\newcommand{\emc}{\em\color{red}}
\newcommand{\Ga}{\Gamma}
\newcommand{\ki}{\raisebox{.8mm}{$\chi $}}
\newcommand{\lj}{\vspace*{2mm}\\}
\newcommand{\Lq}{\lq\lq}
\newcommand{\lra}{\longrightarrow}
\newcommand{\mbb}[1]{\mathbbm{#1}} 
\newcommand{\mbi}[1]{\boldsymbol{#1}}
\newcommand{\n}{\{1,\dots ,n\}}
\newcommand{\ovl}[1]{\overline{#1}}
\newcommand{\q}{\{1,\dots ,q\}}
\newcommand{\R}{\mathbb{R}} 
\newcommand{\Rq}{\rq\rq}
\newcommand{\rr}{\, \mid \,}
\newcommand{\sA}{{\script A}}
\newcommand{\script}[1]{\EuScript{#1}}
\newcommand{\sG}{{\script G}}
\newcommand{\smileq}{\stackrel{\stackrel{q}{\smile}}{\cdots}}
\newcommand{\sP}{{\script P}}
\newcommand{\sr}{\stackrel} 
\newcommand{\tl}{\triangleleft} 
\newcommand{\va}{\emptyset} 

\newtheorem{Theorem}{Theorem}
\newtheorem{Proposition}{Proposition}
\newtheorem{Corollary}{Corollary}

\begin{document}

\thispagestyle{empty}
\vspace*{-30mm}
\begin{center}

{\bf Polyhedrons and Perceptrons Are Functionally Equivalent}\vspace*{1mm}

Daniel Crespin \\ Facultad de Ciencias \\ Universidad Central de
Venezuela\\ \end{center}

\begin{abstract}

\setlength{\parskip}{1.5mm}
\noindent Mathematical definitions of polyhedrons and perceptron networks are discussed. The formalization of polyhedrons is done in a rather traditional way. For networks, previously proposed systems are developed. Perceptron networks in disjunctive normal form (DNF) and conjunctive normal forms (CNF) are introduced. The main theme is that single output perceptron neural networks and characteristic functions of polyhedrons are one and the same class of functions. A rigorous formulation and proof that three layers suffice is obtained. The various constructions and results are among several steps required for algorithms that replace incremental and statistical learning with more efficient, direct and exact geometric methods for calculation of perceptron architecture and weights.

\end{abstract}

\setlength{\parskip}{0mm}
\tableofcontents

\setlength{\parskip}{2mm}

\section{Introduction\label{Intrdctn}}

A perceptron unit is the characteristic function of a linear half space, a perceptron layer is a product of units and a perceptron network is a composable sequence of perceptron layers.

A collection $H_1, \dots, H_n$ of half spaces in $\R^m$ generates a finite Boolean algebra of subsets of $\R^m$. A polyhedron $K$ is any member of this algebra.

With well structured mathematical theories of polyhedrons and perceptron networks a natural proof can be given that {\em for any polyhedron $K$ in $\R^m$ there exists an $m$-input, single output perceptron neural network $P$} which is functionally equivalent to $K$. Functionally equivalent means that the characteristic function $\ki[K]$ is equal to the function $F[P]$, so
\[
\forall K\ \ \exists P \mbox{ such that } \ki[K]=F[P]
\]
Function $F[P]$ is the composition of the layers of $P$, traditional and operationally called a \Lq forward pass\Rq; see section \ref{NtwrkFnctns}. So, from a functional viewpoint {\em all polyhedrons are perceptrons}.

If $K$ is given, calculation of $P$ requires writing $K$ in disjunctive normal form (DNF) and, depending on the way $K$ is specified, at this point intractability may arise. But if polyhedron $K$ is already expressed in disjunctive normal form, $K=K_\DNF[\mbi{H};\De]$, then calculation of the architecture, weights and transfer functions of the network $P$ is immediate. Actually, $P$ can be taken equal to a DNF network, $P=P_\DNF[\mbi{H};\De]$, which can be trivially obtained from scheme $\De$. Schemes are defined in section \ref{Indxs} and DNF polyhedrons in section \ref{DNFPlhdrns}.

The converse is also proved, namely, that {\em all perceptrons are functionally equivalent to polyhedrons}. Given an $m$-input, $k$-layer (arbitrary $k\geq 1$), single output perceptron neural network $P$, there exists a polyhedron $K$ in $\R^m$ which is functionally equivalent to $P$
\[
\forall P\ \ \exists K \mbox{ such that } F[P]=\ki[K]
\]
The existence of $K$ implies that ---rudely stated--- {\em perceptrons cannot do more than polyhedrons}. Generally speaking polyhedron $K$ is calculable from the architecture and weights of perceptron network $P$, but at computational level intractability appears. However, intractability disappears if $P$ is a DNF network.

A DNF perceptron network is a three layer network having conjunctive second layer and disjunctive third layer. See definition in section \ref{DNFNtwrks} below. In case network $P$ is such DNF perceptron network, $P=P_\DNF[\mbi{H};\De]$, there is no intractability to calculate a functionally equivalent polyhedron $K$. In fact, calculation of $K$ as a DNF polyhedron is immediate. That DNF polyhedrons and networks are freely convertible is fortunate. Similarly for their duals, the CNF polyhedrons and CNF networks; see definitions in sections \ref{DNFPlhdrns} and \ref{CNFNtwrks} below.

Our discussion proves in particular that for any perceptron network $P$ there exists a functionally equivalent DNF perceptron network $P_\DNF[\mbi{H};\De]$, that is, such that $F[P]=F[P_\DNF[\mbi{H};\De]]$. This means that {\em for perceptron networks $3$ layers suffice}. Again, when passing from general $P$ to more manageable $P_\DNF[\mbi{H};\De]$ intractability arises. When dealing with data recognition problems it is possible, and always advisable, stay within the realm of DNF polyhedrons and DNF networks. Or within its dual CNF realm.

Replacing DNF polyhedrons with CNF polyhedrons (=intersections of unions of linear half spaces) produces valid dual statements. And furthermore, all results generalize to $r$-tuples of polyhedrons and $r$-output networks, to be detailedly discussed in \cite{Crespin1}.

As mentioned, several results are hindered by intractability. It is a fact, however, that direct and efficient calculation of architecture and weights of DNF perceptron networks that perfectly recognize given data ---and maintains margins, preset at will up to largest theoretically admissible values--- is computationally bland.

\section{Half spaces\label{HlfSpcs}}
A {\emc linear form} is a non-constant function $f:\R^n\to \R$, $f(y_1,\dots , y_n)=w_0+w_1 y_1+\cdots +w_n y_n$. Numerical specification of $f$ is done by the coefficients $w_i$. For $f$ to be non-constant it is necessary and sufficient that at least one of $w_1, \dots, w_m$ be non-zero.

A {\emc half space} is a subset $H$ of $\R^m$ defined by means of inequalities imposed on $f$. Given $f$ two linear inequalities will be considered, namely the {\emc lax inequality}, $f\geq 0$, and the {\emc strict inequality} $f>0$, which define two corresponding half spaces. These are the {\emc lax half space of $f$} or {\emc closed half space of $f$}, denoted $H[f;\geq]$, and the {\emc strict half space of $f$} or {\emc open half space of $f$}, denoted $H[f;>]$ 
\[
H[f;\geq]=\{y\in \R^n \rr f(y)\geq 0\}\ \ \ \ \ \ H[f;>]=\{y\in \R^n \rr f(y)> 0\}
\]
A {\emc half space of $f$} is either $H=H[f;\geq]$ or $H=H[f;>]$. For computational purposes half spaces are specified by the coefficients $w_i$ of the form and one of the inequality symbols $\geq$, or $>$.

Taking complements interchanges lax and strict, or equivalently interchanges open and closed
\[
\R^m-H[f;\geq]=H[(-f);>]\ \ \ \ \ \ \R^m-H[f;>] = H[(-f);\geq]
\]

To begin providing geometric structure, half spaces will be accommodated in $n$-tuples
\[
\mbi{H}=(H_1, \dots, H_n)
\]
In order to specify a subcollection of $\mbi{H}$ it suffices to give an index set $I\con \mbb{n}=\n$. When considering multilayer perceptron networks, use will be made of sequences $\mbi{H}^{(1)}, \dots, \mbi{H}^{(k)}$, with $\mbi{H}^{(i)}$ an $n_i$-tuple of half spaces in $\R^{n_{i-1}}$. See section \ref{PrcptrnNtwrks}.

\section{Cells and cocells\label{CllsAndCclls}}
A cell over $\mbi{H}$ is an intersection of some of the half spaces appearing in $\mbi{H}$, and some of their their complements. To make this precise, take two subsets $I^1,I^0$ of $\mbb{n}$, let $\Ga=(I^1,I^0)$ and define the {\emc cell of $\Ga$ over $\mbi{H}$} as the following intersection of half spaces
\[
C_*[\mbi{H};\Ga]=\bigcap_{i\in I^1}H_i\ \cap\ \bigcap_{i\in I^0}(\R^m - H_i)
\]

If $\mbi{H}$ and $\Ga$ are explicitly known then $C_*[\mbi{H};\Ga]$ is computationally {\emc bland} in the sense that establishing whether or not a numerically given $x\in \R^m$ belongs to the cell is a routine evaluation of linear functions, inequalities and conjunctions, without dreaded complexity issues.

Cells are always convex. They can be bounded or not, and open, closed or neither. In particular, cells are allowed to be non-compact. Half spaces of $\mbi{H}$, and their complementary half spaces, are cells over $\mbi{H}$. For appropriated $\mbi{H}$, the convex hull of finitely many points is a cell. By definition, $d$-dimensional simplexes are convex hulls of $d+1$ affinely independent points, hence simplexes are cells. Simplexes in $\R^m$ of dimension $d<m$ have empty interior, therefore cells can have empty interior. Hyperplanes can be expressed as intersections of two closed half spaces, $H[f;\geq]\cap H[-f;\geq]$. Affine subspaces $A$ are intersections of hyperplanes, hence of half spaces, thus they are cells, with empty interior except when $A=\R^m$.

Dually, the {\emc cocell of $\Ga$ over $\mbi{H}$} is the union
\[
C^*[\mbi{H};\Ga]=\bigcup_{i\in I^1}H_i\ \cup\ \bigcup_{i\in I^0}(\R^m - H_i)
\]

\section{Polyhedrons\label{Plhdrns}}
The {\emc polyhedral algebra generated by the half spaces of $\mbi{H}$} is denoted $\sA_*[\mbi{H}]$ and has a standard existential definition as the smallest collection of subsets of $\R^m$ that includes the ambient space $\R^m$, and the half spaces of $\mbi{H}$, and is closed under intersections and complements. %This is also the smallest collection that contains $\R^m$, contains all possible cells $C_*[\mbi{H};\Ga]$ over $\mbi{H}$ (intersections of half spaces) and is closed under unions.

One can also existentially define the {\emc copolyhedral algebra generated by the half spaces of $\mbi{H}$} as the smallest collection $\sA^*[\mbi{H}]$ containing $\R^m$, the half spaces of $\mbi{H}$, and closed under unions and complements. %The algebra $\sA^*[\mbi{H}]$ is also the smallest collection that contains $\R^m$ and all possible cocells $C^*[\mbi{H};\Ga]$ (unions of half spaces) and is closed under intersections.

As is well known, these are one and the same algebra, $\sA_*[\mbi{H}]=\sA^*[\mbi{H}]$, to be simply denoted $\sA[\mbi{H}]$. The algebra $\sA[\mbi{H}]$ is a finite Boolean algebra of subsets of $\R^m$. Compare with Halmos \cite{Halmos1}, Chapter 1. For a more detailed discussion of Boolean algebras see Halmos and Gehring \cite{Halmos2}.

A {\emc polyhedron over $\mbi{H}$} is by definition a member $K$ of the polyhedral algebra, $K\in \sA_*[\mbi{H}]$. Also by definition, a {\emc copolyhedron over $\mbi{H}$} is any member of the copolyhedral algebra $\sA^*[\mbi{H}]$. And since the algebras are equal the terms polyhedron and copolyhedron designate the same objects. The definitions do not require, and do not provide, an explicit description of $K$ in terms of the $H_i$s. Preference for one of the terms polyhedron/copolyhedron, may depend on intention to allude one of the disjunctive/conjunctive normal forms. See section \ref{DNFPlhdrns} below. It often suffices to mention only polyhedrons.

\section{Variety of polyhedrons\label{VrtyOfPlhdrns}}
Polyhedrons are not required to be convex, nor connected, nor simply connected. Any \Lq higher connectivity\Rq\ can occur. For a more technical statement we invoke standard Algebraic Topology spellings, not to be  conjured elsewhere in this paper. Let $X$ be any finite CW complex of dimension $k$. There are finite simplicial complexes $S$, of same dimension $k$, with geometric realization $|S|$ which is homotopy equivalent to $X$; see \cite{Hatcher}, Chapter 2, Section 2C, Theorem 2C.5. Then, there is a linear embedding of $|S|$ into $\R^m$ with $m= 2k+1$; see Spanier, \cite{Spanier}, Chapter 3, Theorem 9. The image $K$ of the linear embedding is a finite union of simplexes, and because simplexes are cells, $K$ is a polyhedron. Hence polyhedrons $K$ exist with homotopy types of arbitrary finite CW complexes $X$. Perceptron networks $P$ exist with \Lq forward pass\Rq\ function equal to the characteristic function of $K$, $F[P]=\ki[K]$; see Corollary \ref{Crllry1} below. Existence of networks $P$ with characteristic set $K=F[P]^{-1}(1)$ equal to a polyhedron with the homotopy type of an arbitrary finite CW complex, reflects the rich non-linear nature of multilayer perceptron neural networks. On the other hand, being unions of convex cells, polyhedrons are conceptually simple and provide fruitful geometric imagery that is the key for the practical solution of data recognition problems.

\section{Perceptron units\label{PrcptrnUnts}}
Let $H$ be a half space. The {\emc perceptron unit of $H$}, $p[H]$, is its characteristic function
\[
p[H] = \ki[H]
\]

Let $f$ be a linear form. Define then the {\emc lax perceptron unit of $f$}, denoted $p[f;\geq]$, and the {\emc strict perceptron unit of $f$}, denoted $p[f;>]$, as the characteristic functions of the respective half spaces
\[
p[f;\geq]=\ki[H[f;\geq]]\ \ \ \ \ \ p[f;>]=\ki[H[f;>]]
\]

For a given $n$-tuple of half spaces $\mbi{H}=(H_1, \dots,H_n)$ a {\emc perceptron unit of $\mbi{H}$} is a characteristic function of a component half space $H_i$ of $\mbi{H}$
\[
p[H_i]=\ki[H_i]:\R^m\to \B=\{1,0\}
\]

Since $H_i=\ki[H_i]^{-1}(1)$ and $\R^m-H_i=\ki[H_i]^{-1}(0)$, both $H_i$ and $\R^m-H_i$ can be specified by means of $p[H_i]$.

\section{Perceptron layers\label{PrcptrnLyrs}}
The {\emc perceptron layer of $\mbi{H}$} is the product of all the perceptron units of $\mbi{H}$
\[
\mbi{p}[\mbi{H}]= (p[H_1], \dots, p[H_n]):\R^m\to \B^n
\]

If $I=\{i_1, \dots , i_t\}\con \n$ is an index set, the {\emc perceptron layer of $I$ over $\mbi{H}$} is
\[
\mbi{p}[\mbi{H};I]= (p[H_{i_1}], \dots, p[H_{i_t}]):\R^m\to \B^t
\]

As defined, units and layers are functions, bit valued in the first case, binary vector valued in the second. Since $\B\con \R$ and $\B^n\con \R^n$, the codomains of unit $p[H_i]$ and of layer $\mbi{p}[\mbi{H}]$ can be enlarged to make them real valued and real vector valued functions, $p[H_i]:\R^m\to \R$ and $\mbi{p}[\mbi{H}]:\R^m\to \R^n$ respectively.

\section{Perceptron networks\label{PrcptrnNtwrks}}
Consider a sequence $f_i:X_{i-1}\to Y_i$, $i=1, \dots, k$, of functions between sets. The sequence is {\emc composable} if $Y_i\con X_i$, $i=1,\dots, k-1$. In such case there is a well defined composition $f=f_k\comp \cdots \comp f_1:X_0\to Y_k$.

A {\emc $k$-layer perceptron network} is a composable sequence $P=(\mbi{p}^{(1)},\dots, \mbi{p}^{(k)})$ of perceptron layers. The {\emc diagram of $P$} is
\[
\R^{n_0} \sr{\textstyle \mbi{p}^{(1)}}{\lra} \R^{n_1}\sr{\textstyle \mbi{p}^{(2)}}{\lra} \cdots \lra
\R^{n_{k-1}}\sr{\textstyle \mbi{p}^{(k)}}\lra  \B^{n_k}
\]

The {\emc architecture of $P$} is the sequence of dimensions
\[
n_0 \ \cdots \ n_k
\]
Intentionally, this definition of architecture does not keep track of which weights have zero value.

The network is {\emc over $\mbi{H}$} if the first layer is $\mbi{p}[\mbi{H}]$, $\mbi{p}^{(1)}=\mbi{p}[\mbi{H}]$. And is a {\emc single output network} if $n_k=1$. The collection of perceptron networks over $\mbi{H}$ will be denoted $\sP[\mbi{H}]$. All $k$-layer networks, with any $k\geq 1$, are included in $\sP[\mbi{H}]$.

\section{Network functions\label{NtwrkFnctns}}
For $k\geq 2$ perceptron networks are not functions, but composition of their layers are functions. The {\emc function of $P$}, or {\emc perceptron map}, usually referred to as \Lq forward pass of $P$\Rq, is the composition
\[
F[P] = \mbi{p}^{(k)}\comp \mbi{p}^{(k-1)}\comp \cdots\comp \mbi{p}^{(1)}: \R^{n_0} \to \B
\] 

A terminological remark: Let $X$ be any set. The {\emc characteristic function} of a subset $A\con X$ is the bit valued function $\ki[A]:X\to \B$. And the {\emc characteristic set} of a bit valued function $f:X\to \B$ is, by definition, $\ki[f]=f^{-1}(1)$. Note that symbol $\ki$ has duplicity: It is an operator from subsets to bit valued functions. And is also an operator from bit valued functions to subsets. Formalities would require to write something like $\ki_{\rm S}^{\rm F}:{\rm Subsets}\to {\rm Functions}$ and $\ki_{\rm F}^{\rm S}:{\rm Functions}\to {\rm Subsets}$. But there is scant margin for confusion if we simply state that $f=\ki[\ki[f]]$ and $A=\ki[\ki[A]]$.

In particular since $F[P]:\R^{n_0}\to \B$ is bit valued, it is the characteristic function of its characteristic set: $F[P]=\ki[K]$, with $K=F[P]^{-1}(1)$. We now prove that $K$ is a polyhedron.

Simplify notation letting $m=n_0$, $n=n_1$, $\mbi{H}=(H_1, \dots, H_n)$, $\ki_i=\ki[H_i]$ and $\mbi{p}^{(1)}=\mbi{p}[\mbi{H}]$. Write $F[P]=g\comp \mbi{\ki[H]}$ with $g=\mbi{p}^{(k)}\comp \mbi{p}^{(k-1)}\comp \cdots\comp \mbi{p}^{(2)}$. Then each $b=(b_1, \dots, b_n)\in \B^n=\B^{n_1}$ has inverse image by first layer equal to
\[
\mbi{\ki[H]}^{-1}(b) = \bigcap_{i=1}^n \ki_i^{-1}(b_i)
\]
But $\ki_i^{-1}(1)=H_i$ and $\ki_i^{-1}(0)=\R^m-H_i$. Hence $\mbi{\ki[H]}^{-1}(b)$ is a cell over $\mbi{H}$. Taking $B=g^{-1}(1)$ we have
\[
F[P]^{-1}(1) = \bigcup_{b\in B}\mbi{\ki[H]}^{-1}(b)
\]
Thus, the characteristic set of $F[P]$ is a union of cells, hence is a polyhedron. The reasoning only required elementary notions about sets and functions. But the result is crucial for perceptron networks, hence it will be stated as

\begin{Theorem}\label{Thrm0}
The function of a single output network $P$ over $\mbi{H}$ is the characteristic function of some polyhedron over $\mbi{H}$
\[
P\in \sP[\mbi{H}]  \ \ \Rightarrow  \ \  \exists K \in\sA[\mbi{H}] \mbox{ such that } F[P]= \ki[K]
\]
\end{Theorem}

{\em Proof:} Done.

\section{Indexes\label{Indxs}}
To handle DNF/CNF polyhedrons and networks over an $n$-tuple $\mbi{H}$ of half spaces, index sets will be used. Recall that acronym DNF stands for for {\emc disjunctive normal form} and CNF for {\emc conjunctive normal form}.

An index set $I$ is {\emc over $n$} if $I\con \n$. When not empty, set $I$ can always be written as $I=\{i_1, \dots, i_t\}$ with $1\leq i_1 < \cdots < i_t \leq n$. The collection of index sets over $n$ is the power set $2^{\mbb{n}}$ of $\mbb{n}$, which is partially ordered under inclusion. There is also a total order  on $2^{\mbb{n}}$, the lexicographic order, where for $I'=\{i'_1, \dots, i'_{t'}\}$ we put $I'\prec I$ if $i'_j=i_j$ for $j=1, \dots k-1$, and $i'_k < i_k$. In consequence for any collection of index sets, $\mbi{I}=\{I_1, \dots, I_q\}$, it can be assumed that $I_1 \prec \cdots\prec I_q$.

An {\emc index pair over $n$} is a pair $\Ga=(I^1,I^0)$ with components that are index sets over $n$. The index pair is {\emc consistent} if $I^1\cap I^0=\va$. As explained above, index pairs will allow to specify cells over $\mbi{H}$. The collection of index pairs is $2^{\mbb{n}}\times 2^{\mbb{n}}$. Here, with $\Ga'=(I^{'\, 1},I^{'\, 0}$, the lexicographic order is $\Ga'\tl \Ga$ whenever $I^{'\, 1}\prec I^1$, or if $I^{'\, 1}= I^1$ and $I^{'\, 0}\prec I^0$.

An {\emc index pair collection of multiplicity $q$ over $n$} of $\mbi{\Ga}$ is a set $\mbi{\Ga}$ with $q$ elements such that each element i an index pair over $n$
\[
\mbi{\Ga}=\{\Ga_1, \dots, \Ga_q\}=\{(I_1^1,I_1^0),\dots, (I_q^1,I_q^0)\}
\]
It can be assumed that $\Ga_1\tl \cdots \tl \Ga_q$. The collection of all such $\mbi{\Ga}$ constitute a set denoted $\sG^q[n]$,
\[
\sG^q[n]=(2^{\mbb{n}}\times 2^{\mbb{n}})\times \smileq \times (2^{\mbb{n}}\times 2^{\mbb{n}})
\]

A {\emc scheme over $n$} is a pair $\De=(\mbi{\Ga},J)$ with $\mbi{\Ga}\in \sG^q[n]$ and $J\con \q$. Schemes $\De$ will be used to specify DNF and CNF polyhedrons over $\mbi{H}$. They will also define DNF and CNF networks.

\section{Adders\label{Addrs}}
For $I=\{i_1, \dots, i_t\}\con \mbb{n}$ define the {\emc adder of $I$} as the linear form $\Add[I]:\R^n\to \R$ given by
\bea{\label{Addr}}
\ba{rcl}
\Add[I](y) & = & \Add[I](y_1, \dots, y_n)\lj 
           & = & \sum_{i\in I}y_i\lj 
           & = & y_{i_i}+ \cdots + y_{i_t}
\ea
\eea
For binary vectors $b=(b_1,\dots, b_n)\in \B^n$ we have 
\[
\ba{lcl}
\hspace*{8mm}\Add[I](b)=0                  &  \Leftrightarrow  &  \forall i\in I\ b_i=0 \lj
\hspace*{8mm}\Add[I](b)=|I|                &  \Leftrightarrow  &  \forall i\in I\ b_i=1 \lj
0< \Add[I](b)< |I|  &  \Leftrightarrow  &  \mbox{otherwise}
\ea
\]
Here $|I|$ is the number of elements of $I$.

\section{Conjunctive forms\label{CnjnctvFrms}}
Let $I\con \n$. The {\emc original conjunctive linear form of $I$} is the linear form $\Cn[I]:\R^n\to \R$ defined as
\[
\Cn[I] = \Add[I]-|I|+\DS\frac12 
\]
On binary vectors $b\in \B^n$ the values of $\Cn[I]$ are half integers, are {\em never} a whole integer, and by inspection we conclude that
\[
\ba{rcl}
\Cn[I](b) = \ \ \, \DS\frac12  &  \Leftrightarrow  &  \forall i\in I\ b_i=1 \lj 
\Cn[I](b) \leq -\DS\frac12  &  \Leftrightarrow     &  \exists i\in I\ b_i=0 
\ea
\]

The {\emc complementary conjunctive linear form of $I$} is
\[
\ovl{\Cn}[I] = -\Add[I]+\DS\frac12 
\]
Hence for binary vectors
\[
\ba{rcl}
\ovl{\Cn}[I](b) = \ \ \, \DS\frac12  & \Leftrightarrow  & \forall i\in I\ b_i=0 \lj 
\ovl{\Cn}[I](b)  \leq  -\DS\frac12   & \Leftrightarrow  & \exists i\in I\ b_i=1
\ea
\]

Consider a consistent pair $\Ga=(I^1,I^0)$ of index sets over $n$. The {\emc conjunctive linear form of $\Ga$} is
\[
\Cn[\Ga] = \Add[I^1]-|I^1|+\DS\frac12 -\Add[I^0]
\]

Note that, for $I\con \mbb{n}$, $\Cn[I]=\Cn[(I,\va)]$ and $\ovl{\Cn}[I]=\Cn[(\va,I)]$.

On binary vectors $b\in \B^n$ form $\Cn[\Ga]$ satisfies
\[
\ba{rcl}
\Cn[\Ga](b) =\ \ \, \DS\frac12 & \Leftrightarrow  & \forall i\in I^1\ b_i=1 \mbox{ and } \forall i\in I^0\ b_i=0\lj 
\Cn[\Ga](b) \leq  -\DS\frac12  & \Leftrightarrow  & \exists i\in I^1\ b_i=0 \ \mbox{ or }\ \exists i\in I^0\ b_i=1
\ea
\]

\section{Disjunctive forms\label{DsjnctvFrms}}
Define the {\emc original disjunctive linear form of $I$} as the function $\Ds[I]:\R^n\to \R$ given by
\[
\Ds[I] = \Add[I]-\DS\frac12 
\]
which for binary vectors $b\in \B^n$ has values
\[
\ba{rcl}
\Ds[I](b) \geq \ \ \, \DS\frac12  &  \Leftrightarrow  &  \exists i\in I\ b_i=1 \lj 
\Ds[I](b) = -\DS\frac12  &  \Leftrightarrow  &  \forall i\in I\ b_i=0 
\ea
\]

Define also the {\emc complementary disjunctive linear form of $I$} by the expression
\[
\ovl{\Ds}[I] = -\Add[I]+|I| - \DS\frac12 
\]
Evaluated on binary vectors this gives
\[
\ba{rcl}
\ovl{\Ds}[I](b) = \ \ \,  \DS\frac12       & \Leftrightarrow  &  \forall i\in I\ b_i=1 \lj 
\ovl{\Ds}[I](b)  \leq  -\DS\frac12 & \Leftrightarrow  & \exists i\in I\ b_i=0
\ea
\]

Consider again a consistent pair $\Ga=(I^1,I^0)$ of index sets over $\mbb{n}$. The {\emc disjunctive linear form of $\Ga$} is $\Ds[\Ga]:\R^n\to \R$ given by
\[
\Ds[\Ga] = \Add[I^1]-\Add[I^0] +|I^0|-\DS\frac12
\]

Note that, whenever $I\con \mbb{n}$, $\Ds[I]=\Ds[(I,\va)]$ and $\ovl{\Ds}[I]=\Ds[(\va,I)]$. Values on binary vectors $b\in \B^n$ are given by
\[
\ba{rcl}
\Ds[\Ga](b) =  - \DS\frac12     & \Leftrightarrow  & \forall i\in I^1\ b_i=0 \mbox{ and } \forall i\in I^0\ b_i=1\lj 
\Ds[\Ga](b) \geq \ \ \, \DS\frac12  & \Leftrightarrow  & \exists i\in I^1\ b_i=1 \ \mbox{ or }\ \exists i\in I^0\ b_i=0
\ea
\]

\section{Conjunctive units\label{CnjnctvUnts}}
Perceptron units of linear forms $f$ were defined in section \ref{PrcptrnUnts}. For conjunctive linear forms, $f=\Cn[\Ga]$, we obtain the {\emc conjunctive lax perceptron unit of $\Ga$} and the {\emc conjunctive strict perceptron unit of $\Ga$} defined as 
\[
p[\Cn[\Ga];\geq] \ \ \ \ \ \ p[\Cn[\Ga];>]
\]
These are bit valued functions defined on $\R^n$. In the case of conjunctive and disjunctive forms, both lax and strict units will serve our goals equally well. For the sake of definiteness \Lq conjunctive perceptron unit\Rq\ will refer to the lax unit, to be simply denoted $\Ga^\cap$, thus $\Ga^\cap=p[\Cn[\Ga];\geq]$.

For binary vectors we then have
\[
\ba{rcl}
\Ga^\cap(b) = 1  & \Leftrightarrow  & \forall i\in I^1\ b_i=1 \mbox{ and } \forall i\in I^0\ b_i=0\lj 
\Ga^\cap(b) = 0  & \Leftrightarrow  & \exists i\in I^1\ b_i=0 \ \mbox{ or }\ \exists i\in I^0\ b_i=1
\ea
\]

Therefore the characteristic function of cell $C_*[\mbi{H};\Ga]$ is
\[
\ki[C_*[\mbi{H};\Ga]] = \Ga^\cap\comp \mbi{p}[\mbi{H}]
\]

\section{Conjunctive layers\label{CnjnctvLyrs}}
Let $\mbi{\Ga}=\{\Ga_1,\dots, \Ga_q\}$ be a collection of index set pairs over $n$, $\Ga_j=(I_j^1,I_j^0)$. The {\emc conjunctive perceptron layer of $\mbi{\Ga}$}, denoted $\mbi{\Ga^\cap}:\R^n\to \B^q$, is the product of the conjunctive units of its member index set pairs
\[
\mbi{\Ga^\cap} = (\Ga_1^\cap,\dots, \Ga_q^\cap)
\]

The value of $\mbi{\Ga^\cap}$ on a binary vector $b\in \B^n$ has bit components $\Ga_1^\cap(b),\dots, \Ga_q^\cap(b)$. In consequence the $q$-tuple of characteristic functions of the cells is the composition of the perceptron layer of $\mbi{H}$ with the conjunctive perceptron layer of $\mbi{\Ga}$
\[
(\ki[C_*[\mbi{H};\Ga_1]], \dots, \ki[C_*[\mbi{H};\Ga_q]]) = \mbi{\Ga^\cap}\comp \mbi{p}[\mbi{H}]
\]

\section{Disjunctive units\label{DsjnctvUnts}}
The disjunctive linear form $f=\Ds[\Ga]:\R^n\to \R$ defines a {\emc disjunctive lax perceptron unit of $\Ga$} and a {\emc disjunctive strict perceptron unit of $\Ga$}
\[
p[\Ds[\Ga];\geq]:\R^n\to \B \ \ \ \ \ \ p[\Ds[\Ga];>]:\R^n\to \B
\]
Again for definiteness, $\Ga^\cup$ will denote a lax unit, $\Ga^\cup=p[\Ds[\Ga];\geq]$.

The values of $\Ga^\cup$ on binary vectors are
\[
\ba{rcl}
\Ga^\cup(b) = 1  & \Leftrightarrow  & \exists i\in I^1\ b_i=1 \ \mbox{ or }\ \exists i\in I^0\ b_i=0 \lj 
\Ga^\cup(b) = 0  & \Leftrightarrow  & \forall i\in I^1\ b_i=0 \mbox{ and } \forall i\in I^0\ b_i=1
\ea
\]

Hence the characteristic function of the cocell $C^*[\mbi{H};\Ga]$ is
\[
\ki[C^*[\mbi{H};\Ga]] = \Ga^\cup\comp \mbi{p}[\mbi{H}]
\]

\section{Disjunctive layers\label{DsjnctvLyrs}}
As before, let $\mbi{\Ga}=\{\Ga_1,\dots, \Ga_q\}$ be a $q$-tuple of index set pairs over $n$. The {\emc disjunctive perceptron layer of $\mbi{\Ga}$}, denoted $\mbi{\Ga^\cup}:\R^n\to \B^q$, is the product of respective disjunctive units
\[
\mbi{\Ga^\cup} = (\Ga_1^\cup,\dots, \Ga_q^\cup)
\]

Layer $\mbi{\Ga^\cup}$ evaluated on a binary vector $b\in \B^n$ is a binary vector with components $\Ga_1^\cup(b),\dots, \Ga_q^\cup(b)$. Dually to the case of cells, the $q$-tuple of characteristic functions of the cocells is equal to the composition of the layer of $\mbi{H}$ with the disjunctive perceptron layer of $\mbi{\Ga}$
\[
(\ki[C^*[\mbi{H};\Ga_1]], \dots, \ki[C^*[\mbi{H};\Ga_q]]) = \mbi{\Ga^\cup}\comp \mbi{p}[\mbi{H}]
\]

\section{DNF and CNF polyhedrons\label{DNFPlhdrns}}
Let $\De=(\mbi{\Ga},J)$ be a scheme over $n$. By definition the {\emc DNF polyhedron of $\De$ over $\mbi{H}$} is the union of the cells specified by index pairs $\Ga_j$ with $j\in J$
\[
K_\DNF[\mbi{H};\De]=\bigcup_{j\in J}C_*[\mbi{H};\Ga_j]
\]
Polyhedron $K_\DNF[\mbi{H};\De]$ is specified by a of collection $\mbi{H}$ of half spaces; by a list $\mbi{\Ga}=\{\Ga_1, \dots , \Ga_q\}$ of index pairs over $n$ that define cells over $\mbi{H}$; and by an index set $J$ over $q$ that tells which of the cells to include in the union. So defined, DNF polyhedrons are polyhedrons, $K_\DNF[\mbi{H};\De]\in \sA[\mbi{H}]$, endowed with an explicit description.

Define the {\emc CNF polyhedron of $\De$ over $\mbi{H}$} as
\[
K_\CNF[\mbi{H};\De]=\bigcap_{j\in J}C^*[\mbi{H};\Ga_j]
\]
The specification of $K_\CNF[\mbi{H};\De]$ consists of the collection $\mbi{H}$ of half spaces; of a list $\mbi{\Ga}=\{\Ga_1, \dots ,\Ga_q\}$ of index pairs over $n$ with each pair defining a cocell; and of a collection of cocells to be intersected, indicated by the index set $J$ over $q$. This notion is dual of DNF polyhedron. Note that \Lq CNF copolyhedron\Rq\ could have been used as a consistent name for what has been called CNF polyhedron.

When a DNF polyhedron is given, some half spaces of $\mbi{H}$ could eventually be \Lq mute\Rq\ in the sense that they will appear in none of the cells. And some cells may also turn out be mute since they may be left out of the polyhedron. This seems wasteful. But note that different cells are made from different half spaces, and different polyhedrons are made from different cells. Thus, when considering a polyhedron, what for one cell is a mute half space may be needed for another cell. If several polyhedrons are simultaneously discussed, what are mute cells for one of these may be needed for the others. On the other hand and for efficiency, half spaces and cells with participation in more than one object need only appear once. These comments also apply to the CNF case.

\section{DNF and CNF polyhedral algebras\label{DNFPlhdrlAlbr}}
Consider $K\in \sA[\mbi{H}]$. A {\emc DNF presentation of $K$} is a scheme $\De$ such that $K=K_\DNF[\mbi{H};\De]$. The {\emc DNF polyhedral algebra of $\mbi{H}$}, denoted $\sA_\DNF[\mbi{H}]$, is the class of polyhedrons that have some DNF presentation. 

A {\emc CNF presentation of $K$} is a scheme $\De$ such that $K=K_\CNF[\mbi{H};\De]$. The {\emc DNF polyhedral algebra of $\mbi{H}$}, denoted $\sA_\DNF[\mbi{H}]$, is the class of polyhedrons that have some DNF presentation.

That $\sA_\DNF[\mbi{H}]$ and $\sA_\CNF[\mbi{H}]$ (by definition {\em subsets} of $\sA[\mbi{H}]$) are in fact Boolean algebras requires proof. Schemes for unions, intersections and complements have to be calculated in terms of initially given schemes. We now state formally

\begin{Proposition}\label{PrpDNFAlgbr}
$\sA_\DNF[\mbi{H}]$ and $\sA_\CNF[\mbi{H}]$ are Boolean Algebras
\end{Proposition}

{\em Proof:} Elementary. For details see \cite{Crespin0}.

\section{Equality of algebras\label{EqltyBlnAlgbrs}}
The algebra $\sA[\mbi{H}]$ of subsets of $\R^m$ was defined as the Boolean algebra generated by the half spaces of $\mbi{H}$.

Let $\Ga=(\{i\},\va)$, $\Ga_1=\Ga$, $\mbi{\Ga}=\{\Ga_1\}$ and $J=\{1\}$, then scheme $\De=(\mbi{\Ga},J)$ is a DNF presentation over $\mbi{H}$ of the half space $H_i$, that is, $H_i=K_\DNF[\mbi{H};\De]$. Therefore the half space $H_i$ belongs to the DNF polyhedral algebra of $\mbi{H}$, $H_i\in \sA_\DNF[\mbi{H}]$. Also, the same scheme is a CNF presentation over $\mbi{H}$ of $H_i$, \vspace*{1mm}$H_i=K_\CNF[\mbi{H};\De]$, and $H_i\in \sA_\CNF[\mbi{H}]$. Note that taking\vspace*{1mm} $\Ga=(\va, \{i\})$, we similarly obtain a scheme $\ovl{\De}$ for the complementary half spaces, $\R^m - H_i=K_\DNF[\mbi{H};\ovl{\De}]$ and $\R^m - H_i=K_\CNF[\mbi{H};\ovl{\De}]$.

Proposition \ref{PrpDNFAlgbr} implies now that $\sA_\DNF[\mbi{H}]=\sA[\mbi{H}]$ and $\sA_\CNF[\mbi{H}]=\sA[\mbi{H}]$. Therefore

\begin{Theorem}\label{PrpEqltyBlnAlgbrs}
The three Boolean polyhedral algebras are equal
\[
\sA_\DNF[\mbi{H}] = \sA[\mbi{H}] = \sA_\CNF[\mbi{H}]
\]

\end{Theorem}

{\em Proof:} Done.

\section{DNF perceptron networks\label{DNFNtwrks}}
Let $\De=(\mbi{\Ga},J)$ be a scheme over $n$. Define the {\emc DNF perceptron network of $\De$ over $\mbi{H}$} as the three layer, single output perceptron network, denoted $P_\DNF[\mbi{H};\De]$, which has first layer $\mbi{p}[\mbi{H}]$, second layer $\mbi{\Ga^\cap}$ and third layer $J^\cup$, so that $P_\DNF[\mbi{H};\De]=(\mbi{p}[\mbi{H}],\mbi{\Ga^\cap},J^\cup)$. This network has diagram
\[
\R^{m} \sr{\textstyle \mbi{p[H]}}{\lra} \R^n\sr{\textstyle \mbi{\Ga^\cap}}{\lra}
\R^q\sr{\textstyle J^\cup}\lra  \B
\]
Here $\mbi{\Ga^\cap}$ is the conjunctive layer of $\mbi{\Ga}$ defined in section \ref{CnjnctvLyrs}, and layer $J^\cup$ is the disjunctive unit of $J$ described in section \ref{DsjnctvUnts}.

\section{DNF network function\label{DNFNtwrkFnctn}}
Consider a DNF network $P_\DNF[\mbi{H};\De]$ and its network function
\[
F[P_\DNF[\mbi{H};\De]]=J^\cup \comp \mbi{\Ga}^\cap \comp \mbi{p}[\mbi{H}]
\]
From section \ref{CnjnctvLyrs} we know that composition of the first two layers is the product of the characteristic functions of the cells. And from section \ref{DsjnctvUnts} we conclude that further composition with the disjunctive unit of $J$ gives the characteristic function of the DNF polyhedron 
\[
J^\cup \comp \mbi{\Ga}^\cap \comp \mbi{p}[\mbi{H}] = \ki[K_\DNF[\mbi{H};\De]]:\R^m\to \B
\]
This proves, for any $n$-tuple $\mbi{H}$ of half spaces and for any scheme $\De$ over $n$, the following

\begin{Theorem}\label{Thm1}
The DNF polyhedron and the DNF perceptron network of scheme $\De$ over $\mbi{H}$ are functionally equivalent
\[
F[P_\DNF[\mbi{H};\De]] = \ki[K_\DNF[\mbi{H};\De]]
\]
\end{Theorem}

{\em Proof:} Done.

Let $K\in \sA[\mbi{H}]$ be an arbitrary polyhedron over $\mbi{H}$. Theorem \ref{PrpEqltyBlnAlgbrs} proves that some scheme $\De$ exists such that $K=K_\DNF[\mbi{H};\De]$. Theorem \ref{Thm1} gives $\ki[K]=\ki[K_\DNF[\mbi{H};\De]]=P_\DNF[\mbi{H};\De]$ and we reach

\begin{Corollary}\label{Crllry1}

For any polyhedron $K\in \sA[\mbi{H}]$ there exists a functionally equivalent DNF perceptron network $P_\DNF[\mbi{H};\De]$
\[
\ki[K]=F[P_\DNF[\mbi{H};\De]]
\]

\end{Corollary}

{\em Proof:} Done.

\section{CNF perceptron networks\label{CNFNtwrks}}
Dually to section \ref{DNFNtwrks}, the {\emc CNF perceptron network of $\De$ over $\mbi{H}$} is defined as the three layer, single output perceptron network $P_\CNF[\mbi{H};\De]$ with first layer $\mbi{p}[\mbi{H}]$, second layer $\mbi{\Ga}^\cup$, and third layer equal to $J^\cap$, respectively defined in sections \ref{PrcptrnLyrs}, \ref{DsjnctvLyrs} and \ref{CnjnctvUnts}. In symbols, $P_\CNF[\mbi{H};\De]=(\mbi{p}[\mbi{H}],\mbi{\Ga}^\cup,J^\cap)$. The diagram of this perceptron network is
\[
\R^{m} \sr{\textstyle \mbi{p[H]}}{\lra} \R^n\sr{\textstyle \mbi{\Ga^\cup}}{\lra}
\R^q\sr{\textstyle J^\cap}\lra  \B
\]

\section{CNF network function\label{CNFNtwrkFnctn}}
Let $P_\CNF[\mbi{H};\De]$ be a CNF network. Its function is
\[
F[P_\CNF[\mbi{H};\De]] = J^\cap \comp \mbi{\Ga}^\cup \comp \mbi{p}[\mbi{H}]
\]
According to section \ref{DsjnctvLyrs}, composition of the first two layers is the product of the characteristic functions of the cocells. Section \ref{CnjnctvUnts} implies then that composition with the third layer is equal to the characteristic function of the CNF polyhedron
\[
J^\cap \comp \mbi{\Ga}^\cup \comp \mbi{p}[\mbi{H}] = \ki[K_\CNF[\mbi{H};\De]]:\R^m\to \B
\]
Thus, the following dual of Theorem \ref{Thm1} has been proved 

\begin{Theorem}\label{Thm2}
The CNF polyhedron and the CNF perceptron network of scheme $\De$ over $\mbi{H}$ are functionally equivalent.
\[
F[P_\CNF[\mbi{H};\De]] = \ki[K_\CNF[\mbi{H};\De]]
\]
\end{Theorem}

{\em Proof:} Done.

Theorem \ref{PrpEqltyBlnAlgbrs} implies that for any polyhedron $K\in \sA[\mbi{H}]$ there exists CNF presentation $\De$ of $K$, $K=K_\DNF[\mbi{H};\De]$. The dual of Corollary \ref{Crllry1} is

\begin{Corollary}\label{Crllry2}

For any polyhedron $K\in \sA[\mbi{H}]$ there exists a functionally equivalent CNF perceptron network $P_\CNF[\mbi{H};\De]$
\[
\ki[K]=F[P_\CNF[\mbi{H};\De]]
\]

\end{Corollary}

{\em Proof:} Done.

\section{DNF and CNF functional equivalence\label{PlyhAPrcptrnsArEqv}}
Let $P$ be any $m$-input, $k$-layer, single output perceptron network with first layer $\mbi{H}$. Theorems \ref{PrpEqltyBlnAlgbrs}, \ref{Thm1} and \ref{Thm2} allow to conclude the following

\begin{Theorem}\label{Thm3}
There are schemes $\De_*$ and $\De^*$ such that
\[
F[P] = F[P_\DNF[\mbi{H};\De_*]] = \ki(K_\DNF[\mbi{H};\De_*])= F[P_\CNF[\mbi{H};\De^*]] = \ki(K_\CNF[\mbi{H};\De^*])
\]
\end{Theorem}

{\em Proof:} Done.

\section{Three layers suffice\label{ThrLyrsSffc}}
Because DNF ---as well as CNF--- perceptron networks have three layers, taking $P^{(3)}=P_\DNF[\mbi{H};\De_*]$ as immediate consequence of the previous theorem we obtain

\begin{Corollary}\label{CrllryThrLyrs} 
For any $m$-input, $k$-layer, single output perceptron network $P$ with first layer $\mbi{H}$ there exists a functionally equivalent $3$-layer network $P^{(3)}$ over $\mbi{H}$
\[
F[P]=F[P^{(3)}]
\]
\end{Corollary}

The interpretation is that \Lq for perceptrons three layers suffice\Rq, in the precise sense that any function from $\R^m$ to $\B$ realizable by a $k$-layer perceptron network, can also be realized by a network having three layers, and such that for both networks the first layer is the same. See Crespin \cite{Crespin5}.

\section{Conclusions\label{Cnclsns}}
Along the paper polyhedrons and perceptron neural networks have been compared. The context has been one of formal definitions, propositions, theorems and proofs, all within contemporary standards of mathematical rigor. Results were very basic and are natural consequences of definitions. The viewpoint may contribute to establish foundations for a mathematical theory of perceptron neural networks. It is now clear that polyhedrons and perceptron neural networks are functionally the same, P=PNN. So what?

Perceptron networks are often used for pattern recognition. We prefer to talk about {\em data recognition}. Data are finite subsets of $\R^m$. If data sets are given ---non-empty and mutually disjoint--- DNF polyhedrons can be calculated that are adapted to the data, including specification of margins, or distances to \Lq decision boundaries\Rq. Geometry makes possible exquisite adjustments of polyhedrons to data. Conversion of DNF polyhedrons to DNF perceptron networks is immediate, resulting in networks that perfectly recognize the data. Such DNF networks have controllable, ample and flexible generalization capabilities, up to maximum theoretical limits. The methodology has already been software tested. It is considerably simpler and more efficient than backpropagation or support vector machines. The DNF polyhedrons are easy to calculate, and amenable to rule extraction. How to pass from data to polyhedrons will be explained in forthcoming papers.

Direct calculation of DNF polyhedrons provides DNF perceptron networks and competes against learning paradigms. Backpropagation or other types of incremental learning are bypassed. DNF perceptron networks are geometrically gestated and, as in some myths, born with knowledge. The gestation process is brief and efficient. If it is the case that streams of new data keep coming, permanent online gestation would keep the network updated.

That polyhedrons and perceptrons are equivalent is a recurrent theme in neural network literature. The earliest reference known to the present author is the 1987 article \cite{Lippmann} of Lippmann, but older papers may exist. Our own line of development has been circulating in \cite{Crespin3}-\cite{Crespin6}, which papers are available at

http://www.matematica.ciens.ucv.ve/dcrespin/Pub/

and also from

http://ucv.academia.edu/DanielCrespin

\vspace*{12mm}

Oteyeva, Caracas\\
Tuesday, November 05, 2013.

\end{document}